\pdfoutput=1
\documentclass[11pt]{article}
\usepackage[a4paper,margin=2.4cm]{geometry}
\usepackage{graphicx}
\usepackage{xcolor}
\usepackage{amsmath}
\usepackage{booktabs}
\usepackage{longtable}
\usepackage{array}
\usepackage{setspace}
\usepackage[numbers,super,sort&compress]{natbib}

\usepackage{authblk}
\usepackage{caption}
\usepackage[hidelinks]{hyperref}
\hypersetup{pdftitle={Familiar letters, not related languages, let unseen languages into a shared space of meaning},
  pdfauthor={Yuliang Liu, Xinyu Wang, Xiang Bai}}
\title{\textbf{Familiar letters, not related languages, let unseen\\languages into a shared space of meaning}}
\author{Yuliang Liu, Xinyu Wang, Xiang Bai}
\affil{Huazhong University of Science and Technology, Wuhan, China}
\date{\small ylliu@hust.edu.cn}

\begin{document}
\maketitle

\begin{abstract}
\noindent Hundreds of languages have written text but almost no translations, and whether a model learning one from untranslated text builds on a related language or familiar letters is unclear: they usually coincide. Here we show that familiar letters, not a related language, let an unseen language into a shared space of meaning. GlyphMind, a tokenizer-free vision transformer that reads text as pixels and lets any sentence be shown in any script, was pretrained to align translations across 27 writing systems. Adapted on untranslated Lao, a language and script it had never seen, it retrieved English translations seven times as often as the strongest simple baseline. Rewritten in unfamiliar letters, Lao did not enter although its close relative Thai was in pretraining; Santali, which had none, entered in Devanagari, familiar from Hindi, but not in its own script. Transliteration into familiar scripts may thus open multilingual models to languages without translations.
\end{abstract}

Of the more than 7,000 languages spoken in the world, only a small fraction is adequately supported by today's language models\cite{joshi2020state,nllb2024scaling}. These models are typically trained on large multilingual corpora, and prior work has found that they often map translations of the same sentence to nearby points in a shared representation space\cite{pires2019,wu2025semantic}. This sharing is thought to underlie cross-lingual transfer, whereby a model fine-tuned on English examples can perform the same task in other languages\cite{pires2019,conneau2020}. Such models acquire these shared representations either from large amounts of text in each language\cite{conneau2020} or from parallel corpora that pair sentences with their translations\cite{artetxe2019}. However, hundreds of poorly supported languages have only a modest amount of untranslated text\cite{joshi2020state}, and some are written in scripts of their own. It is therefore important to understand how existing multilingual models can be extended to such languages from this limited text alone, without parallel corpora.

Most language models read text through a subword tokenizer, whose vocabulary is learned from their pretraining corpora and therefore favours the languages already in them. As a result, text in rare or unseen languages is split into many more tokens\cite{petrov2023,ahia2023}, and letters absent from the vocabulary may become unknown tokens\cite{pfeiffer2021}. Models perform worse on these languages\cite{ahia2023,pfeiffer2021}, although writing them in the script of a related language can help\cite{muller2021unseen}. Whatever the script, a tokenizer-based model receives text only as numeric codes and never sees what the letters look like.

Pixel models instead read text rendered as images, so they need no tokenizer and accept any script that can be displayed\cite{rust2023pixel,salesky2023multilingual,kesen2025multilingual}. They also see the shapes of letters, and their transfer between languages has been linked to how similar the scripts look\cite{zhang2026seeing}. Yet scripts that look alike often belong to related languages, so this link alone cannot show whether familiar letters themselves help a model adapt. Here we call a script familiar if the model saw it in pretraining or if it looks like a script the model saw. Because a pixel model can read the same sentences in any script, the letters can be changed while the language stays the same.

We developed GlyphMind, a vision transformer that reads each sentence only as a rendered image. We trained it from scratch on sentence pairs in 27 writing systems to place translations close together, so we know which scripts and languages it has read. We then chose two languages absent from pretraining. Lao has a close relative among the training languages, Thai, and the two scripts look alike. We also redrew Lao with the letter shapes of Vai, a script absent from pretraining. Santali has no close relative in training, and its own script, Ol~Chiki, resembles no training script. Santali is also written in Devanagari, which the model reads from Hindi, so we transliterated its text into Devanagari. This gave four versions of two languages, each in familiar and in unfamiliar letters. We first tested whether ordinary Lao could join the shared space after adaptation on about one million untranslated lines. For each of the four versions, we then adapted the model on a matched budget of about 150,000 untranslated lines, mixed with translation pairs replayed from the training languages (Fig.~1). A version entered the shared space if the model then retrieved English translations of its sentences more often than both models adapted on other text and baselines that match sentences by length or ink. In GlyphMind, each language entered the shared space in familiar letters but not in unfamiliar ones. For these two languages, a close linguistic relative was therefore neither necessary nor sufficient for entry. For Santali, the advantage of familiar letters persisted after we removed digits and words shared with the training text. Even without a tokenizer, familiar letters decided whether this much untranslated text was enough.

\begin{figure}[tbp]
\centering
\makebox[\linewidth][c]{\includegraphics[width=180mm]{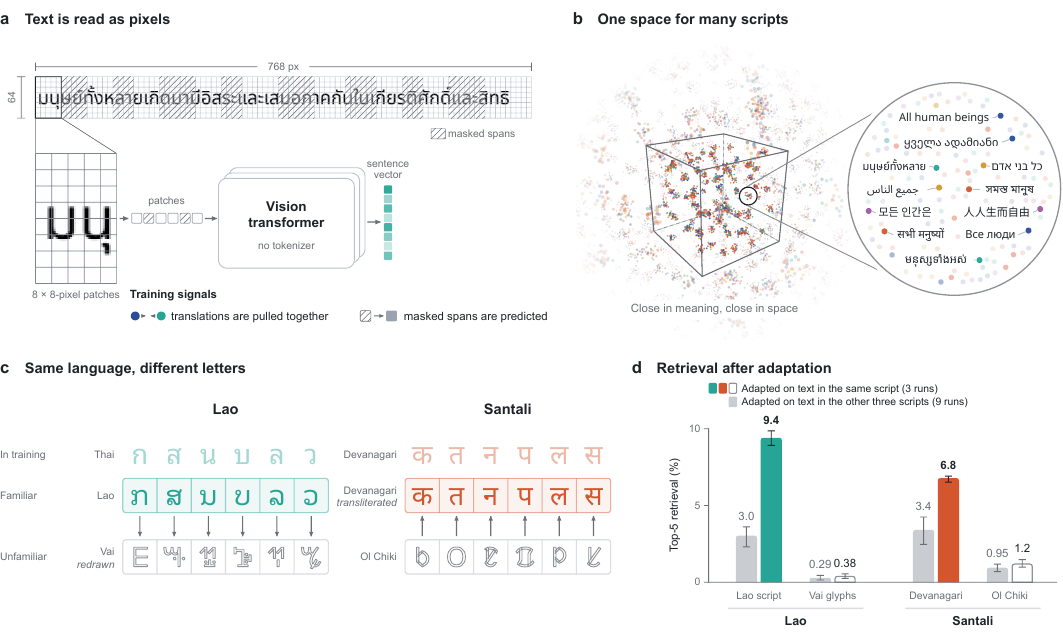}}
\caption{\textbf{Familiar letters let unseen languages into GlyphMind's shared space.} \textbf{a}, GlyphMind reads each sentence as one line image of $64\times768$ pixels, here in Thai. A vision transformer with no tokenizer turns the image's $8\times8$-pixel patches into one sentence vector. In training, translations are pulled together and masked spans are predicted. \textbf{b}, The shared space, drawn schematically. Each dot is a sentence, and translations lie close together. The cube shows a small part of the space. The lens enlarges the cluster for the first sentence of Article~1 of the Universal Declaration of Human Rights. \textbf{c}, Lao letters resemble the Thai letters seen in training. An unfamiliar version redraws each Lao letter as a character of the Vai syllabary. Santali's own script, Ol~Chiki, is absent from training. A familiar version transliterates Santali into Devanagari, which the model reads from Hindi. Santali in Devanagari thus uses the same letters as Hindi, whereas Lao letters only resemble Thai ones. Arrows point from each language's own script to the version made from it. \textbf{d}, Top-5 retrieval of the English translation after adaptation on matched amounts of untranslated text. Lao sentences are ranked against 1,997 candidates and Santali sentences against 1,012, so chance is below 0.5\%. Coloured and hollow bars show three runs adapted on text in the same script as the test sentences. Grey bars show nine runs adapted on text in the other three scripts. Bars show the mean and lines show the range. Colour marks the home region of each script. Blue is Europe, ochre West Asia and Africa, vermilion South Asia, teal Southeast Asia and purple East Asia. Hollow grey marks scripts absent from training.}
\end{figure}

\section*{Results}

GlyphMind reads each sentence as a single line of rendered text, an image of 64 by 768 pixels (Fig.~1a; Extended Data Fig.~1). In pretraining, it learns from scratch to place translations close together, using 4.1 million groups of sentences that translate one another in 27 writing systems (Fig.~1b). Most experiments use GlyphMind-205M, which has 205 million parameters and also learns to predict the features of masked parts of each image (Extended Data Fig.~2a). GlyphMind-38M, with 38 million parameters, learns the translation task alone and is used in the one experiment that trains two models from scratch.

\subsection*{Learning a shared space across scripts}

We first asked whether GlyphMind learns a shared representation space for sentences written in different scripts. Its pretraining sentence pairs come from OPUS\cite{tiedemann2012opus} and Bible translations\cite{akerman2023ebible}, and we tested it on a separate news benchmark, NTREX-128\cite{federmann2022ntrex,barrault2019findings}, which covers 128 languages. For 21 languages in GlyphMind's pretraining, we measured top-1 retrieval accuracy, the proportion of sentences whose English translation ranked first among 1,997 candidates by representation similarity. All 21 use non-Latin scripts, so success cannot rest on letters shared with English.

The median accuracy into English was 30.1\%, against a chance level of 0.05\% (Fig.~2a). Even Armenian, which had only 4,690 lines in pretraining, reached 12.1\% (Fig.~2b). We also compared GlyphMind-205M with two untrained baselines on Russian-to-English retrieval. The trained model achieved 56.5\% accuracy, whereas a randomly initialized encoder and a texture-only network achieved 0.33\% and 0.40\%, respectively (Fig.~2c; Methods).

\begin{figure}[tbp]
\centering
\makebox[\linewidth][c]{\includegraphics[width=180mm]{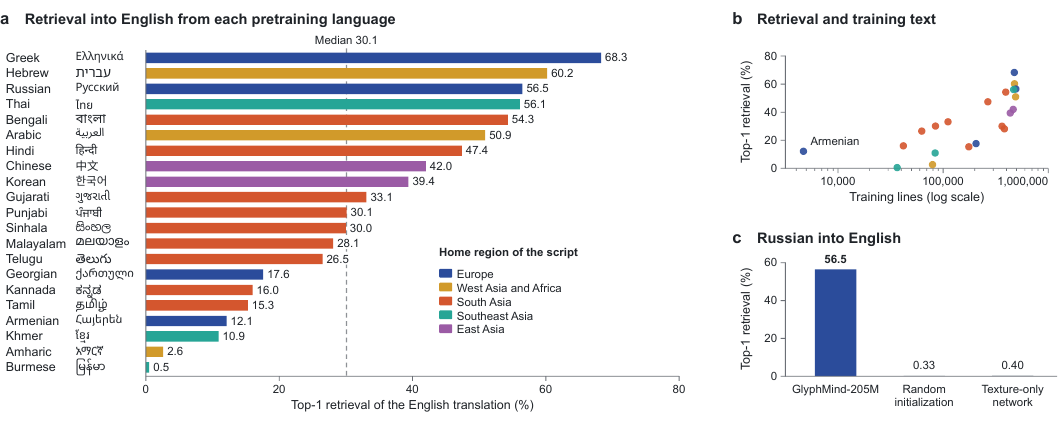}}
\caption{\textbf{GlyphMind's shared space spans many scripts.} \textbf{a}, Retrieval from each pretraining language into English on NTREX-128. Each sentence is compared with 1,997 English sentences. Bars show how often its translation ranks first. They are ordered by accuracy and coloured by the home region of each script, as in Fig.~1. Each language is also named in its own script. The dashed line marks the median of the 21 languages. Chance is 0.05\%. \textbf{b}, Retrieval accuracy from \textbf{a} against the number of pretraining lines in the writing system of each language, on a logarithmic axis (Supplementary Table~1). Armenian had the fewest lines. \textbf{c}, Retrieval from Russian into English for GlyphMind-205M and two untrained networks. One is a copy of GlyphMind-205M, randomly initialized, and the other is a small convolutional network with random weights that sees only image texture.}
\end{figure}

\subsection*{Extending the shared space to an unseen script}

Pretraining had taught GlyphMind-205M to bring images of translated sentences close together. We next asked whether a new language could join this shared space using only untranslated text in that language. We tested this with Lao, whose language and script were both absent from pretraining.

Starting from the pretrained model, we continued training on images of approximately one million untranslated lines of Lao web text, with evaluation sentences removed. The model learned from each Lao image by predicting features of masked spans and of the sentence as a whole. We chose this objective in preliminary experiments (Methods; Extended Data Fig.~2b). Training also continued on translation pairs of the pretraining languages, to preserve the alignment among them.

We evaluated Lao on the same NTREX-128 retrieval task. Because retrieval for a new language starts low, we report top-5 accuracy for new languages. After adaptation, the correct English original ranked among the first five of 1,997 candidates for 14.3\% of Lao sentences, compared with 3.4\% before adaptation and 2.05\% for the strongest simple baseline (Fig.~3a; Extended Data Fig.~3a). NTREX-128 had also guided the choice of adaptation settings, so we tested the same model on 997 Lao-English sentence pairs from the FLORES-200 dev split\cite{nllb2022,goyal2022flores,guzman2019flores}. None of these sentences was in the Lao adaptation text, and the data used for pretraining and replay contained no FLORES-200 sentences. Here the correct translation ranked among the first five for 21.0\% of sentences, compared with 3.0\% for the strongest simple baseline (Fig.~3b). A language absent from pretraining can therefore join the shared space through adaptation on its untranslated text. This came at some cost to a language learned in pretraining: Russian-to-English top-1 retrieval fell from 56.5\% to 44.4\% (Extended Data Table~1).

\begin{figure}[tbp]
\centering
\makebox[\linewidth][c]{\includegraphics[width=180mm]{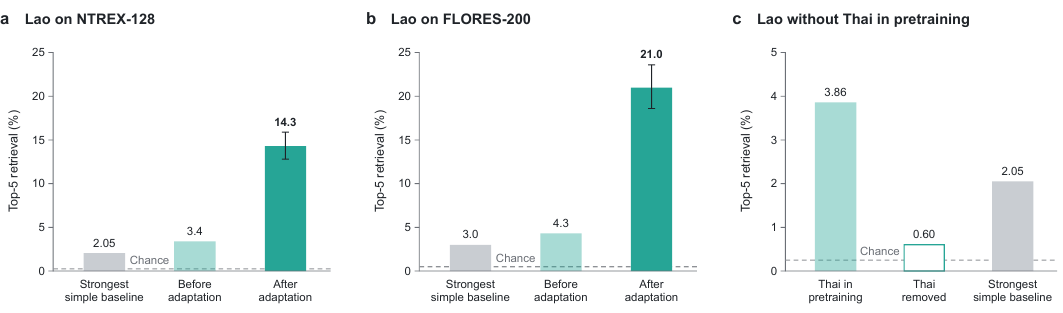}}
\caption{\textbf{A script absent from pretraining enters the shared space.} \textbf{a}, Retrieval from Lao into English on NTREX-128 after adaptation on about one million untranslated Lao lines. Each Lao sentence is compared with 1,997 English sentences. Bars show how often its translation ranks among the first five. The model before adaptation and the strongest of eight simple baselines are shown for comparison. \textbf{b}, The same model on 997 Lao sentences from the FLORES-200 dev split, held out from every choice of adaptation settings. \textbf{c}, Two GlyphMind-38M models pretrained on the same data with and without Thai, scored on NTREX-128 before either saw any Lao text. Each model was trained once. Error bars in \textbf{a} and \textbf{b} are 95\% confidence intervals. Dashed lines mark chance, 0.25\% in \textbf{a} and \textbf{c} and 0.5\% in \textbf{b}.}
\end{figure}

\subsection*{Relatedness and entry into the shared space}

What allowed Lao to enter? One possibility is that a related language or script in pretraining gives adaptation a starting point. Lao has such a relative in Thai, which is close to it in both language and script. We therefore examined whether training on Thai helped Lao retrieval before any Lao adaptation.

We trained two smaller GlyphMind-38M models from scratch in the same way, one on the full pretraining data and one on the same data with all Thai sentences removed. We compared them at the checkpoint where their retrieval on four languages present in both datasets matched (Methods).

Without Thai, Lao top-5 retrieval fell from 3.86\% to 0.60\%, below the 2.05\% of the strongest simple baseline (Fig.~3c; Extended Data Table~2). Thai therefore gave Lao an advantage before the model had seen any Lao text.

\subsection*{The role of script familiarity}

Removing Thai, however, removes both a related language and a related script. To separate the two, we changed the written form of a language while keeping the language itself unchanged. We did this for two languages that differ in whether a close relative was present in pretraining (Fig.~1c).

For Lao, whose close relative Thai was present in pretraining, we compared ordinary Lao with a redrawn version of the same text, in which each Lao character takes the shape of a character from Vai, a script absent from pretraining. Redrawing changed only how the characters looked, leaving the words, word order, grammar and meaning unchanged. The Vai shapes were scaled to fit the original character boxes, so line widths and line breaks were preserved.

For Santali, which had no close relative in pretraining, we compared its own Ol~Chiki script, which the model had never seen, with a transliteration into Devanagari, which it had learned from Hindi.

The Santali corpus has 149,810 lines of Ol~Chiki Wikipedia text, far less than the million-line Lao corpus. We therefore subsampled Lao to the same size, which also tested whether this smaller budget was enough for entry. Both versions of each language were made from the same source text, so they carried the same linguistic content.

All four versions were adapted in the same way, on about 150,000 untranslated lines with three random seeds each. Every adapted model was tested on all four versions at a final checkpoint fixed in advance (Extended Data Fig.~3c). On each test set, the three models adapted on that version were compared with the nine adapted on the other three. Because all twelve models were trained further in the same way, an advantage for the three reflects what they learned from that version.

Models adapted on ordinary Lao reached a mean top-5 accuracy of 9.4\% on ordinary Lao (Fig.~1d), against 3.4\% before adaptation and 3.0\% for the models adapted on the other versions. Lao thus still benefited from adaptation on far less text than in the million-line experiment. Redrawn Lao gave no clear benefit. On the redrawn test images, models adapted on redrawn Lao reached 0.38\%, against 0.29\% for the other models and a chance level of 0.25\%. When 500 example translations were used to map the redrawn sentences onto English (Methods; Extended Data Fig.~4), the two groups scored 3.2\% and 3.0\%.

Santali showed the complementary pattern. Models adapted on Santali in Devanagari reached 6.8\% on Devanagari Santali, against 3.4\% for the other models, and every Devanagari run beat all nine others (Extended Data Table~3). Because all models saw the same test images, this lead reflects adaptation rather than the queries being drawn in Devanagari. On Ol~Chiki Santali, models adapted in Ol~Chiki reached only 1.2\%, below the 3.36\% sentence-length baseline. With 500 example translations, they reached 6.4\%, against 3.8\% for the other models (Extended Data Table~3).

Two further Ol~Chiki settings varied the initialization checkpoint and learning rate, and the two strongest of the three settings were extended after an optimizer reset. Across these runs and their checkpoints, retrieval without example translations peaked at 1.78\%, below the 3.36\% sentence-length baseline (Methods and Supplementary Table~3).

Familiarity with the written form therefore decided whether adaptation on untranslated text brought a language into the shared space. A close relative did not help Lao once it was redrawn, and Santali benefited from a familiar script without one.

\subsection*{Testing alternative explanations}

Changing the written form also changed other properties of the text. We examined whether ink, the share of each sentence visible in the image or content shared with the training text could explain the results. We also checked for prior exposure to Santali and to the evaluation sentences.

Redrawn Lao carries about 40\% less ink than ordinary Lao, measured as the total darkness of the pixels, although a template matcher distinguishes its letters as accurately as the originals at the tested sizes. To test whether lighter ink explained the lack of benefit, we drew ordinary Lao with lighter or thinner strokes, or in a lighter font. In all three cases, models adapted on ordinary Lao still scored about two to four times as high as the model before adaptation (Fig.~4a; Extended Data Table~4).

\begin{figure}[tbp]
\centering
\makebox[\linewidth][c]{\includegraphics[width=180mm]{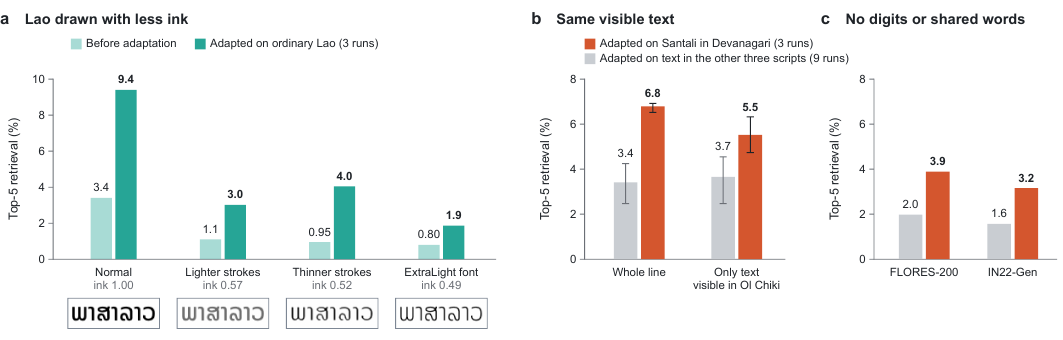}}
\caption{\textbf{The advantage of familiar letters persists in checks on ink, visible text and shared digits and words.} All panels show how often the English translation ranks among the first five, without example translations. \textbf{a}, Ordinary Lao on NTREX-128, drawn with lighter strokes, thinner strokes or a lighter font so that it carries about as little ink as the redrawn Lao. Ink is measured on the test lines relative to normal renders. Below each condition, the first word of a Lao sentence is drawn the same way. \textbf{b}, Santali in Devanagari on FLORES-200, scored on whole lines and on only the text that each line shows in Ol~Chiki. \textbf{c}, Santali in Devanagari with digits and words shared with the pretraining text removed, on FLORES-200 and IN22-Gen. Bars show means over runs, and lines in \textbf{b} show their range.}
\end{figure}

For Santali, Devanagari had two possible advantages besides familiarity. The same text is narrower in Devanagari than in Ol~Chiki, so more of each sentence fits in the fixed-width image before it is cut. Restricting the test images to the text visible in Ol~Chiki kept about half of the advantage (Fig.~4b), although during adaptation the Devanagari lines still showed more of each sentence. Santali in Devanagari also shares digits and words with the pretraining text. Removing them lowered mean top-5 accuracy from 6.8\% to 3.9\%, still above the 2.0\% of the other models on the same filtered test images (Fig.~4c). Visible text and shared digits and words thus each explain part of the advantage, but neither explains all of it.

The advantage could also reflect Santali seen in pretraining. We searched the Devanagari, Bengali and Odia pretraining text for sentences containing several distinctive Santali words. No pretraining sentence passed this screen, although it catches about two-thirds of Santali corpus lines written in Devanagari (Extended Data Table~4). Separately, ten of the twelve four-version runs replayed a mixture that holds the English sides of the FLORES-200 Santali test pairs. We therefore repeated the comparisons on IN22-Gen, whose sentence pairs were absent from all training mixtures. The Devanagari advantage persisted there, including after shared digits and words were removed (Fig.~4c; Extended Data Table~4).

\section*{Discussion}

GlyphMind shows that a model trained on rendered text can extend a shared representation space to an unseen language from untranslated text alone. Whether adaptation succeeded depended on the written form: familiar letters enabled entry even without a close linguistic relative, and unfamiliar letters prevented it despite one. Knowledge of meaning learned from some languages can therefore carry over to another when its written form connects the new input to what the model has already learned.

This matters for models that read text as images. With pixel input, a new script needs no place in a predefined vocabulary, and unfamiliar characters can be shown to the model directly. Yet this does not ensure that the model can use them to reach its existing representations of meaning. Redrawing Lao preserved the language and its meaning but changed the outcome of adaptation. The look of the text therefore shapes what can be learned from the same linguistic evidence. Familiar letter shapes may provide reusable structure through which new sentences become linked to existing representations.

The comparison also separates two sources of transfer that usually occur together. Related languages often share vocabulary and grammar, and related scripts share letter shapes and conventions, so their contributions are hard to tell apart when both change between languages. Changing only the written form within each language showed that a close relative could not make up for unfamiliar letters, whereas familiar letters supported entry without one. The familiar scripts were of two kinds: Devanagari, seen in pretraining, and Lao, unseen but similar to Thai. This suggests that visual experience can support transfer beyond the scripts a model was trained on, and makes script relationships worth studying as a source of transfer.

These findings also bear on language coverage. The resources of a language are usually counted in text, annotations and translations. Our results suggest that the value of those resources also depends on the form in which they reach the model: the same untranslated text can succeed or fail depending on how it is written. Transliteration offers one way to use this. An established transliteration could connect a new language to visual features the model has already learned. The conversion could run inside the model's input pipeline while speakers keep their own script. Input representation thus becomes part of extending multilingual models, alongside collecting data and improving learning objectives.

These findings come from one model family and two adapted languages, yet they open a research direction into how visual experience supports the extension of multilingual knowledge. A central question is which properties of familiar writing systems provide reusable structure for learning a new language. Varying letter shapes step by step while keeping the text fixed could show whether transfer depends on strokes, on whole letters or on broader visual regularities. Using different written forms for adaptation and for testing could show whether familiar letters matter more for learning a language or for reading it afterwards. Such understanding could guide training methods that carry knowledge of meaning across increasingly unfamiliar scripts. Learning correspondences between unfamiliar symbols and known ones offers another route, with possible relevance to historical scripts and decipherment. Extending the evaluation to translation and other tasks would show how far entry into the shared space supports broader language abilities.

Broader language coverage depends on how well new languages can draw on knowledge already acquired from others. Here, familiar letters provided this connection through adaptation on untranslated text. Writing systems thus still matter when text is represented directly as pixels. Treating written form as part of the learning problem opens possibilities for extending multilingual knowledge to languages with little translation data.

\clearpage
\section*{Methods}
\small

\subsection*{Rendering and pretraining data}
\textbf{Rendering.} Each sentence is rendered as one greyscale line image of 64$\times$768 pixels. The font size shrinks until the text fits, down to about 28 pixels. Text that still does not fit is cut at a word or character boundary, and text that overflows the image height is shrunk further. Most evaluation lines are drawn at this smallest size, and many are cut (Extended Data Table~4). We use Noto fonts (Supplementary Tables~1 and~2). In training, each render is shifted and rotated, and Gaussian noise is added. Evaluation renders are clean and deterministic. Training and all evaluations share one renderer, and every image is rendered from digital Unicode text.

\textbf{Pretraining data.} The pretraining mixture contains 4,113,720 meaning groups in 27 writing systems (Supplementary Table~1). A meaning group is a set of sentences that translate one another, and the groups yield 4,606,949 training records. The sentences come from OPUS\citep{tiedemann2012opus}, with English as the pivot language, and from the eBible corpus of verse-aligned Bible translations in native scripts\citep{akerman2023ebible}. A record holds an anchor render, renders of other members of its group as positives, and up to three hard negatives (\emph{Objective}). Each OPUS group gives one record, with its English sentence as the anchor. Each Bible verse gives one record per translation, in which that translation is the anchor and up to eight other translations, drawn at random, are the positives.

\textbf{Decontamination.} The pretraining mixture excludes the FLORES-200 evaluation sets\citep{nllb2022}. Ten adaptation runs replayed from a FLORES-200 variant of this mixture, which also holds the FLORES-200 sentences of 24 languages and has 4,652,213 records (\emph{Four versions of two languages}). The 23-script mixture of the preliminary model also contained the FLORES-200 evaluation sets (\emph{Adaptation to a new language}). Eight of the 1,997 English sentences of NTREX-128\citep{federmann2022ntrex} occur verbatim as lines of the pretraining mixture. They can account for at most 0.4 points of a direct-retrieval score on NTREX-128. Content hashes of every line of the training data will be released (Data and code availability).

\subsection*{Model and pretraining}
\textbf{Architecture.} GlyphMind is a vision transformer\citep{vaswani2017attention,dosovitskiy2021image} with layer normalization before each sublayer. It splits each line image into 8$\times$8-pixel patches (Fig.~1a) and adds fixed two-dimensional sinusoidal position encodings. It has $L$ transformer blocks of width $d$, with feed-forward layers four times wider. Mean pooling after a final layer normalization gives one feature vector per image. A two-layer projection head maps this vector to a $k$-dimensional, L2-normalized embedding. GlyphMind-205M has $d{=}1024$, $L{=}16$, 16 attention heads and $k{=}128$, for 204.9 million parameters. GlyphMind-38M has $d{=}512$, $L{=}12$, 8 heads and $k{=}64$. All models are pretrained from random weights.

\textbf{Objective.} The only supervision is which renders share a meaning. The contrastive loss is InfoNCE\citep{oord2018representation,chen2020simclr} over positives within each batch, with temperature 0.07. Positives are other members of the same meaning group, rendered in other scripts. With probability 0.3, the positive is instead a new substitution cipher of the group's Latin-script text (Extended Data Fig.~1b). Each cipher has fresh letter shapes, which discourages shortcuts specific to one script. A margin loss with margin 0.2 also separates each render from hard negatives (Supplementary Table~2). These negatives shuffle letters within words, replace one word or change digits in the source text before rendering. At each step, one positive and one hard negative are drawn at random from each record. Building these examples uses word boundaries, digits and letter identities, but this information reaches the model only through the images. GlyphMind-205M also has a self-supervised term that predicts the features of masked spans. Its targets come from a teacher that is an exponential moving average of the model\citep{baevski2022data2vec,grill2020bootstrap}. The spans cover about half of the ink (Supplementary Table~2).

\textbf{Optimization.} We trained GlyphMind-205M with AdamW on a cosine schedule on eight A100 GPUs (Supplementary Table~2). The schedule spans 60 epochs, with a checkpoint every five epochs. We stopped training after epoch 40, when its Russian-to-English top-1 retrieval on NTREX-128 stopped improving, falling from 57.0\% to 56.5\% between checkpoints. This final checkpoint is GlyphMind-205M before adaptation.

\textbf{GlyphMind-38M models.} Two GlyphMind-38M models use the objective above without the predictive term. Supplementary Table~2 gives their optimizer settings. They were trained with and without Thai (\emph{Removing Thai from pretraining}). Extended Data Fig.~2a compares the one with Thai with GlyphMind-205M during pretraining.

\subsection*{Adaptation to a new language}
\textbf{Choice of objective.} We chose the adaptation objective in preliminary experiments with a smaller model (Extended Data Fig.~2b and Supplementary Note~1). This model was pretrained on a 23-script mixture, which contained the FLORES-200 evaluation sets, so the comparisons of objectives are scored on NTREX-128. The objective adopted below predicts masked spans and adds sentence-level targets. Its configuration came from an eight-run sweep. We chose it by the FLORES-200 devtest Lao scores of the final checkpoints. This split is disjoint from FLORES-200 Lao dev, the second Lao test set.

\textbf{Recipe.} Adaptation trains GlyphMind-205M further on untranslated lines of a new language, starting from a pretraining checkpoint. On these lines the model predicts the features of masked spans and two sentence-level targets, the pooled feature and the projection output. The three terms each have weight 1.0, and the mask ratio is 0.5. A frozen copy of the starting checkpoint provides the targets. Translation pairs replayed from a pretraining mixture make up a fixed share of training, the replay fraction, to preserve the alignment learned in pretraining. The learning rate is constant after a short linear warm-up (Supplementary Table~2).

\textbf{Lao.} The Lao corpus has 991,775 lines from CC-100 Lao and Lao Wikipedia. We kept lines of 20--160 characters with at least 60\% of their characters in the Lao Unicode block and removed duplicates. Like every adaptation corpus, it was filtered by exact and shingled substring matching against every protected evaluation sentence, with NTREX-128 as the reference list. The filter removed one NTREX-128 sentence from the CC-100 Lao crawl. Eight configurations ran for 12 epochs each and replayed from the pretraining mixture. They varied the initialization checkpoint, the learning rate and the replay fraction (Supplementary Table~3). Two configurations initialized from epoch 40 then ran for 12 more epochs at the same learning rate, after a reset of the optimizer state. These ten runs form the Lao sweep. The reported Lao run is the extension with a learning rate of $5{\times}10^{-5}$ and a replay fraction of 0.7. Fig.~3a,b and Extended Data Table~1 report this run, and Extended Data Fig.~3a shows its training curve across the reset.

\textbf{Ol~Chiki and subsampled Lao.} The Ol~Chiki corpus has 149,810 deduplicated lines of Santali Wikipedia. We ran three settings of the Lao recipe, each for 80 epochs with a replay fraction of 0.5. This presents about as many lines as 12 epochs of the Lao corpus. The first setting started from epoch 40 with a learning rate of $3{\times}10^{-5}$ and replayed from the pretraining mixture. It is the seed-0 Ol~Chiki run of the four-version experiment. Two additional settings varied the initialization checkpoint and learning rate, and the two strongest runs were extended by 12 epochs after resetting the optimizer (Supplementary Table~3). Direct retrieval stayed below the sentence-length baseline of 3.36\% throughout these runs (\emph{Retrieval evaluation}). A subsampled Lao run used the first Ol~Chiki setting for 80 epochs without a reset, on 149,810 lines drawn at random from the Lao corpus, rendered in the three Lao font families. It is the seed-0 run of ordinary Lao in the four-version experiment.

\subsection*{Retrieval evaluation}
\textbf{Test sets.} We used the 1,997 newstest2019 sentences of NTREX-128\citep{federmann2022ntrex} in US English, Lao and the languages of Fig.~2a, including Russian. No NTREX-128 Lao sentence occurs in any corpus used for pretraining or adaptation. For Santali we used the 1,012 sentences of FLORES-200 devtest\citep{nllb2022}, in Ol~Chiki and transliterated into Devanagari. These Santali sentences never entered pretraining or adaptation, although ten adaptation runs replayed their English translations (\emph{Four versions of two languages}). The second Santali test set is IN22-Gen\citep{gala2023indictrans2}. Its 1,024 Santali sentences in Ol~Chiki and their English translations were collected independently of FLORES-200. None of its Santali sentences occurs whole in the Santali corpora, and none of its English sentences occurs in the pretraining mixture or its FLORES-200 variant. The second Lao test set, FLORES-200 Lao dev, has 997 sentences that occur in no training mixture or Lao adaptation corpus. Their English translations are absent from the pretraining mixture, which also supplied the replay of the Lao sweep.

\textbf{Pure renders.} Queries in scripts absent from pretraining are drawn as pure renders, which keep only spaces and the characters of the script's own Unicode block. This removes Latin fragments, punctuation and shared symbols, but keeps the script's own digits. The blocks are U+0E80--0EFF for Lao and U+1C50--1C7F for Ol~Chiki. Santali in Devanagari is transliterated from the reduced Ol~Chiki text. Queries in pretraining languages use unmodified renders, except for the pure renders of Thai (U+0E00--0E7F) and Khmer (U+1780--17FF) in Extended Data Table~2.

\textbf{Retrieval.} Each query is compared with English candidates by the cosine similarity of their pooled backbone features. Top-$k$ accuracy is the fraction of queries whose translation ranks among the first $k$ candidates. We report top-5 unless stated otherwise, so chance is five divided by the number of candidates. Direct retrieval uses no translations of the new language and ranks each query against every English sentence of the test set. In retrieval with example translations, 500 test pairs map the new language onto English (Extended Data Fig.~4b). The pooled backbone features are centred and projected onto a 64-dimensional principal subspace for each language, fitted without pairing information on sentences outside the test set. An orthogonal Procrustes rotation\citep{schonemann1966generalized} fitted on the 500 pairs maps one subspace onto the other, and candidates are scored by dot product. Queries and candidates are then restricted to held-out pairs disjoint from the 500, because a rotation can memorize its fitting pairs (Extended Data Fig.~4c). These number 800 for NTREX-128, 500 for FLORES-200 Santali and IN22-Gen, and 400 for FLORES-200 Lao dev. Results for the reported Lao run on NTREX-128 and in Extended Data Table~1 average three random draws of the 500 pairs. The four-version experiment, the scores on FLORES-200 Lao dev and the checks on alternative explanations average ten.

\textbf{Baselines and controls.} Eight simple baselines ignore meaning. Four compare the length of the raw or pure text, by absolute difference or by rank, and the others use ink area or word count, each by absolute difference or by rank. Direct retrieval on Lao is compared with the strongest simple baseline. On NTREX-128 this baseline matches raw length by absolute difference and reaches 2.05\%. For Santali, the sentence-length baseline is the strongest of the four length baselines on the Ol~Chiki text. It reaches 3.36\%, higher than for Lao, because line lengths in an alphabetic script track English closely. The matched Devanagari test set (\emph{Four versions of two languages}) is compared only between models. Retrieval with example translations is compared with chance and with two controls (Extended Data Fig.~4a). One control is a randomly initialized GlyphMind-205M. In the other, the rotation is fitted on example translations paired at random. Both stay near chance for Lao. Fig.~2c compares GlyphMind-205M with two untrained baselines. One is a second random initialization of GlyphMind-205M. The other is a frozen texture-only network with 52,096 random parameters. Two convolutional layers (32 and 64 channels, 5 by 5 kernels, stride 2, each followed by a ReLU) and global average pooling turn the image into a 64-dimensional embedding.

\textbf{Geometry agreement.} Extended Data Tables~1 and~2 and Extended Data Fig.~2 also report geometry agreement, which asks whether two languages' sentences are arranged alike. It is the Spearman correlation between the upper triangles of their cosine-similarity matrices\citep{kriegeskorte2008representational}. Each matrix is first residualized linearly against its own length-similarity matrix. A sentence's length is the number of Unicode characters in the text it is rendered from, and the length similarity of two sentences is the negative absolute difference of their lengths. Because a randomly initialized GlyphMind-205M already scores 0.184 for Lao, agreement is compared only between models.

\subsection*{Removing Thai from pretraining}
We trained two GlyphMind-38M models from scratch with identical code, recipe and schedule, one on the full pretraining mixture and one without its 468,940 Thai renders. Both mixtures contain the same 4,113,720 meaning groups, and the 26 other scripts have identical counts. Records anchored on Thai were removed with it, so the model without Thai saw 8.1\% fewer records and took 8.1\% fewer optimizer steps per epoch. Because further training erodes what a model carries over to Lao from related scripts, this asymmetry favours the model without Thai. Each model was trained for 30 epochs. After epoch~10, both runs continued with a fresh optimizer state and a new 20-epoch cosine schedule (Supplementary Table~2).

At epochs 10, 20 and 30, we measured Thai-to-English retrieval to check the removal, and Lao direct retrieval and geometry agreement on pure renders (Extended Data Table~2). Agreement was also measured for Khmer, a related script in both mixtures, and for Ol~Chiki, which has no relative in either. As fixed in advance, the models were compared where their trained-script skill matched. This skill is the mean top-1 retrieval for Russian, Greek, Hindi and Georgian, four languages present in both mixtures. It matched at epoch~10, at 28.5\% with Thai and 28.6\% without. The primary measure fixed in advance was Lao geometry agreement. Fig.~3c shows Lao direct retrieval, a secondary measure (\emph{Statistics and reproducibility}). Lao direct retrieval and pure-render agreement were lower without Thai at all three epochs. Unmodified renders keep digits and Latin fragments. On them, Lao agreement without Thai was 0.027 lower at epoch~10, about a third of the pure-render difference. It was higher without Thai at epochs 20 and 30 (Extended Data Table~2).

\subsection*{Four versions of two languages}
The experiment uses two languages, each in a familiar and an unfamiliar script (Fig.~1c). The four versions are ordinary and redrawn Lao, and Santali in Devanagari and in Ol~Chiki. Both versions of a language are made from the same text. Santali used the 149,810-line Ol~Chiki corpus and its transliteration into Devanagari, and Lao used the subsampled Lao corpus (\emph{Adaptation to a new language}). Each version was adapted three times, with seeds that offset the fixed seed of the training code by 0, 1 or 2. All twelve runs used the first Ol~Chiki setting, starting from the final GlyphMind-205M checkpoint and training for 80 epochs without an optimizer reset. They saved a checkpoint every ten epochs (Extended Data Fig.~3c) and are reported at epoch~80.

\textbf{Replay pools.} Seed~0 of ordinary Lao and of Santali in Ol~Chiki replayed from the pretraining mixture, like the Lao sweep. The other ten runs replayed from its FLORES-200 variant. These are seeds 1 and 2 of every version and seed~0 of redrawn Lao and of Santali in Devanagari. The FLORES-200 variant holds the 2,009 FLORES-200 dev and devtest sentences in each of 24 languages, including English and Hindi, but none in Lao, Santali or Ol~Chiki. The NTREX-128 Lao test sets are unaffected. The FLORES-200 Santali test pairs, however, share their English side with the replayed sentences. We therefore repeated every Santali comparison on IN22-Gen and tested the ten runs for memorized pairs (\emph{Checks on alternative explanations}).

\textbf{Redrawn Lao.} Ordinary Lao is drawn in three Noto font families, Sans, Serif and Looped. For each, we built a font that redraws every Lao letter as a character of the Vai syllabary (Supplementary Note~2). Vai is not among the 27 pretraining scripts, and no training line contains a Vai character. All three fonts take their shapes from one Vai font, so redrawn Lao has less variety of letter style than ordinary Lao. In 818 corpus lines, at least one redrawn font rendered the line exactly like its original font. We removed these lines, leaving 148,992 (Extended Data Table~4).

\textbf{Santali in Devanagari.} Santali is written in several scripts, including Devanagari\citep{everson2005olchiki}. We transliterated the Ol~Chiki corpus and test sets letter by letter into Devanagari with a fixed table that follows Devanagari spelling (Supplementary Note~2). Devanagari is drawn in Noto Sans Devanagari, the font used for it in pretraining, and Ol~Chiki in Noto Sans Ol~Chiki.

\textbf{Scoring.} Every adapted model, the model before adaptation and a randomly initialized GlyphMind-205M were scored on pure renders of four test sets. These are NTREX-128 Lao drawn in Noto Sans Lao and in its redrawn version, and FLORES-200 Santali in Ol~Chiki and in Devanagari. On each test set, the three runs adapted on that version are compared with the nine runs adapted on the other versions (Fig.~1d and Extended Data Table~3). A fifth, matched test set transliterates only the characters that the renderer draws of each Ol~Chiki line, so both scripts show the same words. These shorter lines fit without cutting and are drawn larger for every model.

\subsection*{Checks on alternative explanations}
\textbf{Ink.} We measure ink as the total darkness of a line's pixels. Redrawn Lao lines carry 0.57--0.60 of the ink of ordinary ones (Extended Data Table~4). A template matcher identified isolated letters of ordinary Lao, redrawn Lao and Ol~Chiki with at least 98.9\% accuracy at each of four sizes from 16 to 44 pixels (Supplementary Note~3). We also drew NTREX-128 Lao in Noto Sans Lao with about as little ink as redrawn Lao (Fig.~4a). We lightened every stroke to 0.57 of its darkness, thinned the strokes with a $2{\times}2$ morphological filter, or used the Noto Sans Lao ExtraLight font. In direct retrieval, the three runs adapted on ordinary Lao beat the model before adaptation on every lighter render. On redrawn renders, they scored 0.27\%, against 0.35\% for that model. All trained models scored far lower on the lighter renders than on normal ones, so GlyphMind is sensitive to stroke weight (Extended Data Table~4).

\textbf{Visible text.} Ordinary and redrawn Lao lines are cut at the same place. Santali is about a third narrower in Devanagari, so Devanagari test lines show a median 53\% of their characters, against 34\% in Ol~Chiki (Extended Data Table~4). The matched test set removes this difference at test time. On it, the Devanagari runs retained about half of their 3.4-point lead in direct retrieval over the nine other runs (Fig.~4b and Extended Data Table~3).

\textbf{Digits and shared words.} Santali in Devanagari shares digits with Hindi and some spellings with the pretraining text. We scored every model on Devanagari test lines without digits, which occur in 196 of the 1,012 lines. A second version also removes the words that occur three or more times in the Devanagari lines of the pretraining mixture, its FLORES-200 variant or the Bible corpus, 19\% of all words. In direct retrieval, the lead of the Devanagari runs over the nine other runs fell from 3.4 points on whole lines (2.4--4.3) to 2.1 without digits (1.3--2.9). Without digits or shared words, it was 1.9 points (1.2--2.7), and the Devanagari runs still scored about twice as high as the others (Fig.~4c). With 500 example translations, the lead fell from 4.2 points on whole lines to 3.1 without digits or shared words (1.6--4.6).

\textbf{Prior exposure.} We tested whether Santali written in another script had entered pretraining. A screen for frequent Santali words written in Devanagari, Bengali or Odia passes 67.5\% of the first 20,000 Devanagari lines of the Santali corpus and none of the 1,012 FLORES-200 Hindi sentences (Supplementary Note~3). No pretraining line passed, and none contains an Ol~Chiki character (Extended Data Table~4).

\textbf{Second test set and replay.} On IN22-Gen, transliterated and reduced as above, the Devanagari runs led the other runs in direct retrieval by 3.1 points on whole lines (2.2--4.0). After removing digits and shared words, 12\% of its words, they led by 1.6 points (0.9--2.3). On IN22-Gen in Ol~Chiki, the Ol~Chiki runs gained 0.6 points in direct retrieval over the model before adaptation ($-0.2$ to 1.3). With 500 example translations, they gained 2.0 points ($-0.1$ to 3.9). Memorized pairs would raise scores more on FLORES-200 than on NTREX-128. Compared with the other two runs, the ten runs that replayed FLORES-200 sentences gained 0.5 points in Hindi-to-English direct retrieval on FLORES-200 devtest. On NTREX-128 they gained 0.8 points, a difference of $-0.4$ (95\% interval $-2.4$ to 1.7, Extended Data Table~4).

\subsection*{Statistics and reproducibility}
Single scores carry Wilson 95\% confidence intervals. Differences between groups of models carry 95\% intervals from 2,000 bootstrap resamples of the test lines, paired across models. Each group's models are first averaged line by line, so these intervals exclude seed variation. In direct retrieval, the reported Lao run found the translation in the top five for 285 of 1,997 sentences (95\% interval 12.8--15.9\%). We compared this count with 10,000 random permutations of the English candidates and with 10,000 permutations within sentence-length strata. These gave means of 5.0 and 11.7 hits and maxima of 17 and 27, so $p<10^{-4}$ in both tests. The result was stable over the last four checkpoints (Extended Data Fig.~3a).

\textbf{Seeds and checkpoints.} For each version of the four-version experiment, we report the mean and range of three seeds. Every other configuration was run once, so differences between those models include seed variation. Adaptation results use final checkpoints. Pretrained models are compared at matched skill in \emph{Removing Thai from pretraining}.

\textbf{Selection of the reported Lao run.} Every choice in the Lao sweep used NTREX-128, so the NTREX-128 scores of the reported Lao run may be optimistic. NTREX-128 Russian retrieval also determined when pretraining stopped. We therefore also scored the reported run and the other nine runs of the sweep on FLORES-200 Lao dev, which no choice used (Fig.~3b). The reported Lao run again ranked first, with and without example translations. In direct retrieval it reached 21.0\% (95\% interval 18.6--23.6\%), against 4.3\% before adaptation and 3.0\% for the strongest simple baseline. Its lead over the mean of the other nine runs was 9.5 points (7.5--11.6). On NTREX-128 the lead was 6.2 points (5.0--7.4). With 500 example translations, scored on 400 pairs, the lead on FLORES-200 Lao dev was 8.7 points (5.9--11.6).

\textbf{Analyses fixed in advance.} A dated analysis plan, which we will release (Data and code availability), recorded the following before the runs that tested them. It fixed the rule that an extension is reported at its final checkpoint. It fixed the criteria for Ol~Chiki entry, among them direct retrieval above the simple baselines, which Ol~Chiki did not meet. For the two GlyphMind-38M models of \emph{Removing Thai from pretraining}, it named Lao geometry agreement on pure renders at matched skill as the primary measure. It predicted that this agreement would fall to the Ol~Chiki level without Thai. Without Thai it was 0.157, against 0.230 with Thai and 0.033 for Ol~Chiki (Extended Data Table~2). A planned second stage, adapting both models on Lao and measuring entry, was not run. The design, recipe, scoring checkpoint and reading of the four-version experiment were fixed before ten of its twelve runs began. The other two, the seed-0 runs of ordinary Lao and of Santali in Ol~Chiki, had the same final checkpoint fixed in advance. Entry of redrawn Lao would point to the language, and failure would point to the script. The checks on FLORES-200 Lao dev, ink, shared words, IN22-Gen and replayed FLORES-200 sentences were designed after the main results were known.

\subsection*{Compute}
All runs used one machine with eight NVIDIA A100-80GB GPUs. We count A100-hours as wall time, measured from checkpoint times, multiplied by the number of GPUs used. Pretraining GlyphMind-205M took about 2,100 A100-hours, and the two GlyphMind-38M models of \emph{Removing Thai from pretraining} about 600. Adaptation took about 750 A100-hours for the Lao sweep, 300 for the Ol~Chiki and subsampled Lao runs and 690 for the other ten four-version runs. The preliminary model and its comparison of objectives took about 700. The reported runs total about 5,100 A100-hours, or about 6,200 with exploratory runs.

\subsection*{Use of large language models}
Large language models, mainly Claude (Anthropic) and also ChatGPT (OpenAI), were used as assistants for programming, data processing and analysis, and for polishing the text. The authors conceived and designed the study, decided on and carried out all experiments, verified all results and take full responsibility for the content of this paper.

\subsection*{Data and code availability}
All corpora used are public. They are OPUS\citep{tiedemann2012opus}, the eBible corpus\citep{akerman2023ebible}, Wikipedia dumps (Wikimedia), CC-100, the NTREX-128 test set\citep{federmann2022ntrex,barrault2019findings}, the IN22-Gen benchmark\citep{gala2023indictrans2} and the FLORES-200 benchmark\citep{nllb2022,goyal2022flores,guzman2019flores}. The weights of GlyphMind-205M before and after adaptation on Lao are available at \url{https://huggingface.co/Leon5201314/glyphmind}, with a demo that reproduces the NTREX-128 retrieval results of Fig.~2c and Fig.~3a. On publication, the adaptation corpora, the redrawn Lao fonts with their letter mapping, content hashes of every line of the pretraining mixture and the checkpoints of the four-version experiment will be deposited there; other checkpoints are available from the corresponding authors on request. The pretraining text itself is not redistributed, as some of its sources do not permit it; the test sets are available from their providers. The code for training, adaptation, corpus construction and evaluation will be released on publication under the MIT licence on GitHub, archived at Zenodo, together with the dated analysis plan.

\bibliographystyle{nature-review}
\bibliography{references}

\clearpage
\begin{figure}[p]
\centering
\makebox[\linewidth][c]{\includegraphics[width=180mm]{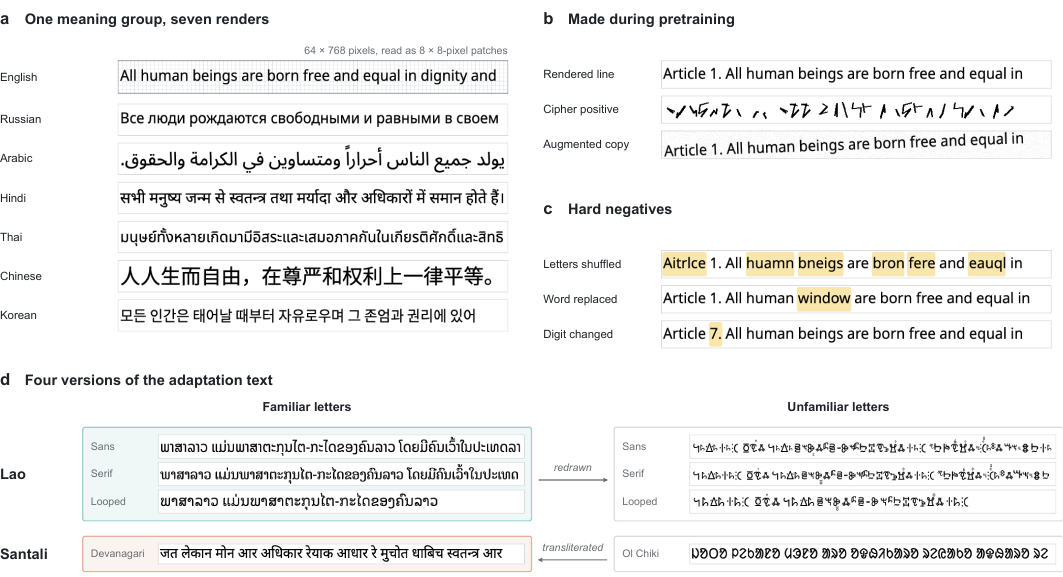}}
\caption*{\textbf{Extended Data Fig.~1 $|$ What GlyphMind is given.} Each row is one model input, a greyscale image of 64 by 768 pixels. The English row in \textbf{a} shows the whole image and its 8 by 8-pixel patches, and the other rows are cropped to the band that holds ink. Text is drawn at the largest size that fits, down to about 28 pixels, and longer text is cut at a word or character boundary. \textbf{a}, The first sentence of Article 1 of the Universal Declaration of Human Rights in seven pretraining languages. Renders of one meaning group serve as positives for one another. \textbf{b}, Variants made from the English line during pretraining. The cipher positive replaces each letter with a new random shape. The augmented copy is shifted, rotated and given noise. \textbf{c}, Hard negatives change the text before it is rendered. Shading marks the changed words. \textbf{d}, One Lao and one Santali sentence in the four versions of the experiment. Ordinary Lao is drawn in three fonts, and each redrawn font replaces the Lao letters with Vai characters. Coloured cells mark familiar letters, as in Fig.~1c. The cipher shapes, noise level, negatives, Vai letters and Devanagari spelling are examples made for this figure.}
\end{figure}

\begin{figure}[p]
\centering
\makebox[\linewidth][c]{\includegraphics[width=180mm]{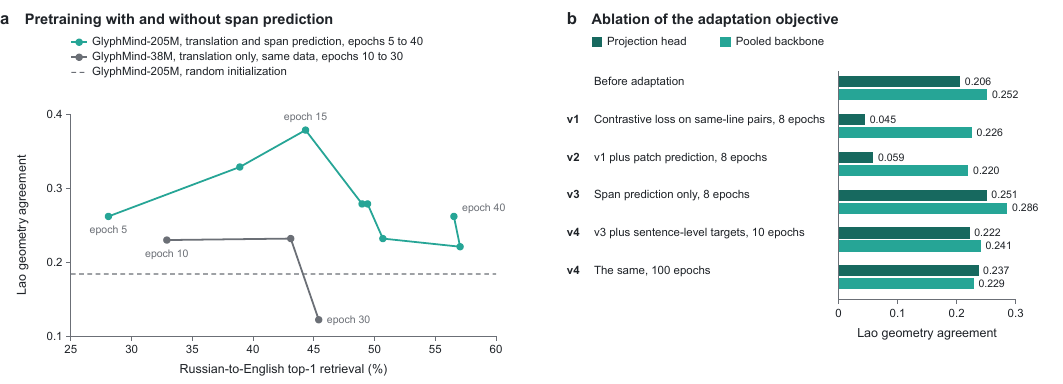}}
\caption*{\textbf{Extended Data Fig.~2 $|$ The objectives of pretraining and adaptation.} \textbf{a}, Lao-English geometry agreement before any Lao text, against Russian-to-English top-1 retrieval on NTREX-128, at successive checkpoints of pretraining. GlyphMind-205M learns translation and span prediction. The GlyphMind-38M model trained on the same data with Thai learns translation alone. Agreement uses pure renders and the pooled backbone (Methods). The two models also differ in size and learning-rate schedule, and checkpoints of GlyphMind-205M five epochs apart differ by up to 0.1. The dashed line is a randomly initialized GlyphMind-205M. \textbf{b}, Adaptation objectives compared in preliminary experiments on a smaller model (Methods). Each run started from the same checkpoint and trained on 48,084 lines of Lao Wikipedia. Agreement is read at the projection head and at the pooled backbone, on NTREX-128 pure renders.}
\end{figure}

\begin{figure}[p]
\centering
\makebox[\linewidth][c]{\includegraphics[width=180mm]{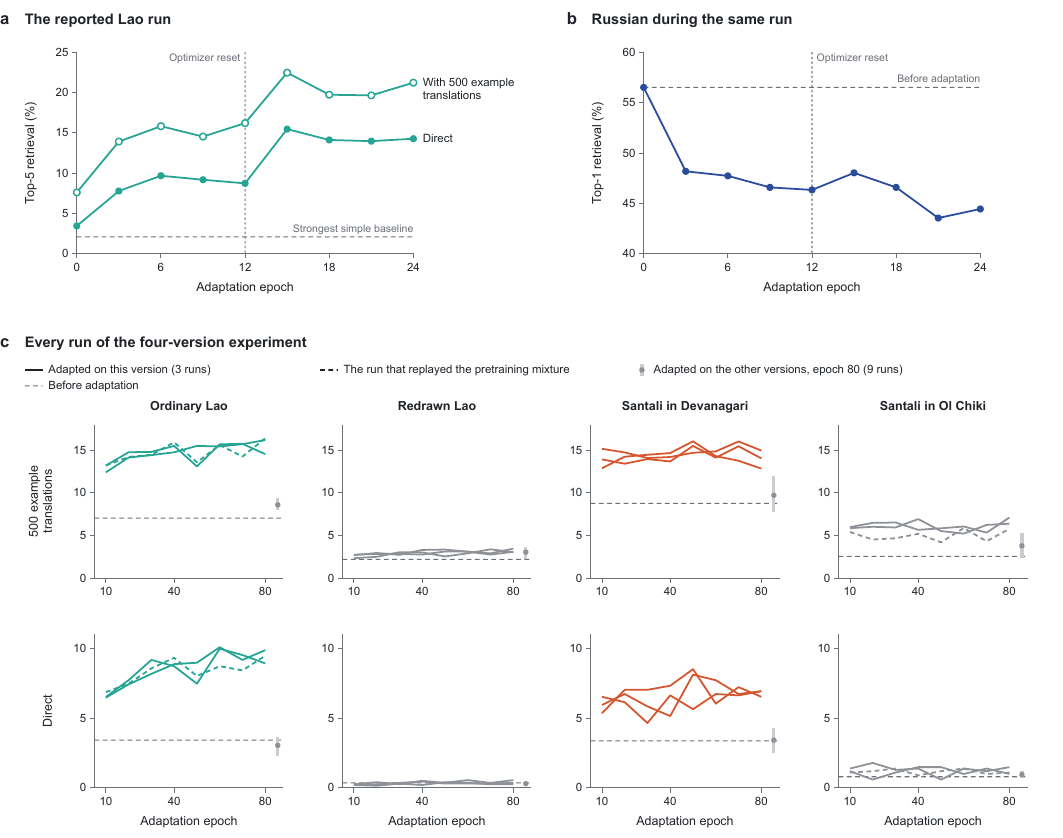}}
\caption*{\textbf{Extended Data Fig.~3 $|$ Training curves of adaptation.} \textbf{a}, The reported Lao run of GlyphMind-205M, scored every three epochs on NTREX-128 pure renders. The optimizer state was reset after epoch 12, and the final checkpoint, epoch 24, is reported. With 500 example translations, scores average three draws of the pairs. \textbf{b}, Russian-to-English top-1 retrieval on unmodified renders during the same run. \textbf{c}, Every run of the four-version experiment, scored every ten epochs on its own test set, NTREX-128 for Lao and FLORES-200 devtest for Santali. Lines show the three runs adapted on each version, and a dashed line marks a run that replayed the pretraining mixture. With 500 example translations, scores average ten draws. The grey bar and dot give the range and mean of the nine runs adapted on the other versions at epoch 80. Dashed horizontal lines show the model before adaptation. Every run is reported at epoch 80, as fixed in advance (Methods).}
\end{figure}

\begin{figure}[p]
\centering
\makebox[\linewidth][c]{\includegraphics[width=180mm]{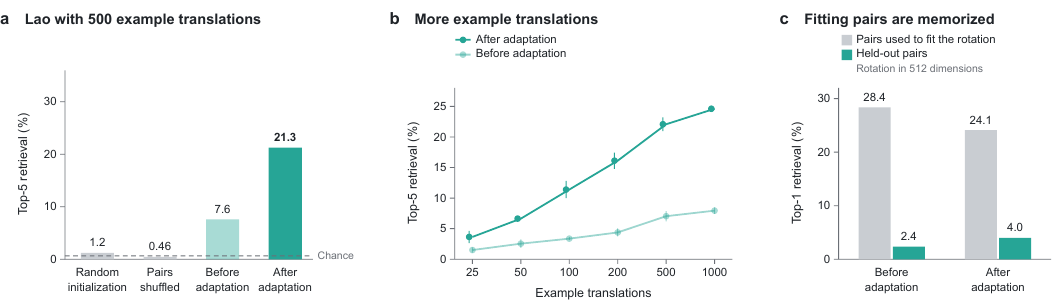}}
\caption*{\textbf{Extended Data Fig.~4 $|$ Retrieval with example translations.} \textbf{a}, Lao-to-English retrieval of the reported Lao run on NTREX-128 with 500 example translations, scored on 800 held-out pairs and averaged over three draws. The controls are a randomly initialized GlyphMind-205M and example translations paired at random. \textbf{b}, The same measure against the number of example translations, for GlyphMind-205M before and after adaptation, as mean and s.d. over ten draws. \textbf{c}, In the preliminary model, a rotation fitted on about 1,000 pairs in 512 dimensions scores far higher on its own pairs than on held-out pairs.}
\end{figure}

\begin{table}[p]
\centering\footnotesize
\setlength{\tabcolsep}{4pt}
\caption*{\textbf{Extended Data Table~1 $|$ GlyphMind-205M before and after adaptation on Lao.} The rows show GlyphMind-205M before adaptation, after 12 epochs on 991,775 Lao lines, and at the final checkpoint of the reported Lao run, 12 epochs after a reset of the optimizer state. The last row is a randomly initialized GlyphMind-205M. Another random initialization gives 0.33\% Russian retrieval (Fig.~2c). Geometry agreement measures whether Lao and English sentences are arranged alike, with sentence length controlled, on NTREX-128 pure renders (Methods). It is read at the projection-head output and at the pooled backbone. Lao retrieval uses NTREX-128 pure renders. With 500 example translations, scores average three draws of the pairs and rank 800 held-out candidates. Direct retrieval ranks all 1,997 English sentences. Russian retrieval uses unmodified renders. Bold marks the reported Lao run.}
\begin{tabular}{@{}lccccc@{}}
\toprule
 & \multicolumn{2}{c}{Geometry agreement} & \multicolumn{2}{c}{Lao retrieval, top-5 (\%)} &
Russian (\%) \\
\cmidrule(lr){2-3}\cmidrule(lr){4-5}\cmidrule(l){6-6}
Checkpoint & head & backbone & 500 examples & direct & top-1 \\
\midrule
Before adaptation & 0.231 & 0.262 & 7.58 & 3.41 & 56.5 \\
After 12 epochs & 0.242 & 0.241 & 16.21 & 8.71 & 46.3 \\
Final, 12 epochs after the reset & 0.217 & 0.180 & \textbf{21.25} & \textbf{14.27} & 44.4 \\
\midrule
Randomly initialized & 0.181 & 0.184 & 1.21 & 0.65 & 0.25 \\
\bottomrule
\end{tabular}
\end{table}

\begin{table}[p]
\centering\footnotesize
\setlength{\tabcolsep}{4pt}
\caption*{\textbf{Extended Data Table~2 $|$ Removing Thai from pretraining.} Two GlyphMind-38M models were trained on the same mixture, with and without its Thai renders, and scored at epochs 10, 20 and 30 (Methods). They are compared at epoch 10, where their trained-script skill matches. Trained-script skill is the mean top-1 retrieval into English for Russian, Greek, Hindi and Georgian, four languages in both mixtures. All measures use NTREX-128, except that Ol~Chiki uses FLORES-200. Pure renders keep only the characters of a script's own Unicode block. Geometry agreement is read at the pooled backbone unless a row says otherwise. Chance is 0.05\% for top-1 and 0.25\% for top-5 retrieval. The strongest simple baseline reaches 2.05\% in Lao direct retrieval.}
\begin{tabular}{@{}lcccccc@{}}
\toprule
 & \multicolumn{3}{c}{With Thai} & \multicolumn{3}{c}{Without Thai} \\
\cmidrule(lr){2-4}\cmidrule(l){5-7}
Epoch & 10 & 20 & 30 & 10 & 20 & 30 \\
\midrule
\multicolumn{7}{@{}l}{\emph{Retrieval into English (\%)}} \\
Thai, top-1, unmodified renders & 32.3 & 40.2 & 42.0 & 0.55 & 0.60 & 0.30 \\
Thai, top-1, pure renders & 25.5 & 32.2 & 35.0 & 0.20 & 0.25 & 0.15 \\
Trained-script skill, top-1 & 28.5 & 35.7 & 37.5 & 28.6 & 33.9 & 35.5 \\
Lao, direct, top-5, pure renders & 3.86 & 3.46 & 1.55 & 0.60 & 0.65 & 0.60 \\
\midrule
\multicolumn{7}{@{}l}{\emph{Geometry agreement with English}} \\
Lao, pure renders (primary measure) & 0.230 & 0.232 & 0.122 & 0.157 & 0.151 & 0.096 \\
Lao, pure renders, projection head & 0.268 & 0.292 & 0.188 & 0.177 & 0.149 & 0.121 \\
Lao, unmodified renders & 0.328 & 0.244 & 0.185 & 0.301 & 0.306 & 0.218 \\
Khmer, pure renders & 0.283 & 0.288 & 0.247 & 0.263 & 0.344 & 0.242 \\
Ol~Chiki, pure renders & 0.026 & 0.019 & $-$0.005 & 0.033 & 0.012 & 0.018 \\
\bottomrule
\end{tabular}
\end{table}

\begin{table}[p]
\centering\footnotesize
\setlength{\tabcolsep}{3pt}
\caption*{\textbf{Extended Data Table~3 $|$ The four-version experiment, every model on every test set.} Top-5 retrieval into English, in per cent, at epoch 80 of every run. Lao is tested on NTREX-128 and Santali on FLORES-200 devtest. ex, retrieval with 500 example translations, averaged over ten draws and scored on 800 held-out pairs for Lao and 500 for Santali. dir, direct retrieval against all 1,997 English sentences for Lao and 1,012 for Santali. Bold marks the three runs adapted on the version being tested. The matched test set keeps only the characters that each Ol~Chiki test line shows, transliterated into Devanagari (Methods). Chance is 0.63\% for Lao and 1.0\% for Santali with example translations, and 0.25\% and 0.49\% in direct retrieval. The strongest simple baseline in direct retrieval reaches 2.05\% for Lao and 3.36\% for Santali. On their own test set, the three runs average 15.7 (ex) and 9.4 (dir) for ordinary Lao and 3.2 and 0.38 for redrawn Lao. They average 14.0 and 6.8 for Santali in Devanagari, 13.7 and 5.5 on the matched set, and 6.4 and 1.2 for Santali in Ol~Chiki. On the same test sets, the nine runs adapted on the other versions average 8.6 and 3.0, 3.0 and 0.29, 9.7 and 3.4 (9.6 and 3.7 matched), and 3.8 and 0.95. \textsuperscript{a}These runs replayed the pretraining mixture. All other runs replayed its FLORES-200 variant (Methods).}
\resizebox{\linewidth}{!}{%
\begin{tabular}{@{}lcccccccccc@{}}
\toprule
 & \multicolumn{2}{c}{Ordinary Lao} & \multicolumn{2}{c}{Redrawn Lao} &
\multicolumn{2}{c}{Santali, Devanagari} & \multicolumn{2}{c}{Devanagari, matched} &
\multicolumn{2}{c}{Santali, Ol~Chiki} \\
\cmidrule(lr){2-3}\cmidrule(lr){4-5}\cmidrule(lr){6-7}\cmidrule(lr){8-9}\cmidrule(l){10-11}
Model & ex & dir & ex & dir & ex & dir & ex & dir & ex & dir \\
\midrule
GlyphMind-205M before adaptation & 7.03 & 3.41 & 2.20 & 0.35 & 8.76 & 3.36 & 9.96 & 3.46 & 2.56 & 0.79 \\
Randomly initialized GlyphMind-205M & 1.18 & 0.65 & 1.14 & 0.35 & 1.04 & 0.59 & 1.28 & 0.59 & 0.82 & 0.49 \\
\midrule
\multicolumn{11}{@{}l}{\emph{Adapted on ordinary Lao}} \\
\ seed 0\textsuperscript{a} & \textbf{16.40} & \textbf{9.46} & 3.63 & 0.25 & 9.86 & 3.66 & 9.78 & 4.15 & 3.82 & 0.89 \\
\ seed 1 & \textbf{16.19} & \textbf{9.86} & 3.06 & 0.40 & 11.96 & 4.25 & 10.48 & 4.55 & 2.38 & 0.69 \\
\ seed 2 & \textbf{14.54} & \textbf{8.91} & 3.43 & 0.15 & 9.54 & 3.46 & 9.48 & 4.05 & 4.02 & 0.89 \\
\multicolumn{11}{@{}l}{\emph{Adapted on redrawn Lao}} \\
\ seed 0 & 9.40 & 2.65 & \textbf{3.06} & \textbf{0.35} & 9.12 & 2.47 & 9.58 & 3.26 & 3.90 & 0.89 \\
\ seed 1 & 8.06 & 2.30 & \textbf{3.08} & \textbf{0.25} & 9.90 & 3.06 & 10.14 & 2.47 & 4.36 & 0.89 \\
\ seed 2 & 7.94 & 2.80 & \textbf{3.46} & \textbf{0.55} & 10.72 & 4.05 & 9.76 & 3.95 & 3.58 & 1.19 \\
\multicolumn{11}{@{}l}{\emph{Adapted on Santali in Devanagari}} \\
\ seed 0 & 8.82 & 2.75 & 3.28 & 0.30 & \textbf{12.84} & \textbf{6.52} & \textbf{13.90} & \textbf{5.53} & 5.28 & 0.89 \\
\ seed 1 & 9.12 & 3.56 & 2.23 & 0.35 & \textbf{14.04} & \textbf{6.92} & \textbf{12.86} & \textbf{6.32} & 3.38 & 1.19 \\
\ seed 2 & 8.66 & 3.61 & 3.21 & 0.20 & \textbf{14.96} & \textbf{6.92} & \textbf{14.20} & \textbf{4.74} & 3.36 & 0.99 \\
\multicolumn{11}{@{}l}{\emph{Adapted on Santali in Ol~Chiki}} \\
\ seed 0\textsuperscript{a} & 8.41 & 3.20 & 2.69 & 0.45 & 8.54 & 3.16 & 9.86 & 3.46 & \textbf{5.78} & \textbf{1.09} \\
\ seed 1 & 8.81 & 3.36 & 2.75 & 0.35 & 10.06 & 3.36 & 9.26 & 4.25 & \textbf{7.10} & \textbf{1.48} \\
\ seed 2 & 8.16 & 3.15 & 3.11 & 0.20 & 7.74 & 3.26 & 7.60 & 2.77 & \textbf{6.38} & \textbf{0.99} \\
\bottomrule
\end{tabular}}
\end{table}

\begin{table}[p]
\centering\footnotesize
\setlength{\tabcolsep}{4pt}
\caption*{\textbf{Extended Data Table~4 $|$ Checks on alternative explanations.} Each block reports one check described in Methods. Lao is drawn in Noto Sans Lao unless a row names another font. Retrieval is into English, top-5 and direct unless a row says otherwise. In the ink block, the three values are the mean of the three ordinary-Lao runs, the model before adaptation and the reported Lao run. In the Devanagari block and the Ol~Chiki row, they are the mean of the three runs adapted on that version, the mean of the nine other runs and the model before adaptation. Gains are over the model before adaptation. Shared words occur three or more times in the Devanagari lines of the pretraining mixture, its FLORES-200 variant or the Bible corpus. Leads are in percentage points, with 95\% intervals from 2,000 bootstrap resamples of the test lines, paired across models (Methods).}
\begin{tabular}{@{}p{0.36\linewidth}p{0.60\linewidth}@{}}
\toprule
Check & Result \\
\midrule
\multicolumn{2}{@{}l}{\emph{How test lines are drawn}} \\
NTREX-128 Lao & median font size 28\,px, 90\% at 40\,px or less. 73\% of lines cut, and a line
shows a median 56\% of its characters \\
FLORES-200 Santali, Ol~Chiki & 28\,px. 100\% of lines cut, and a line shows 34\% of its
characters \\
FLORES-200 Santali, Devanagari & 28\,px. 94\% of lines cut, and a line shows 53\% of its
characters \\
Matched Devanagari test set & lead over the nine other runs 1.9 points, with all three Devanagari runs above all nine others, against 3.4 on whole lines \\
\midrule
\multicolumn{2}{@{}l}{\emph{Letters identified by a template matcher, 30 noisy copies of each
(\%, ordinary / redrawn Lao)}} \\
44\,px & Sans 99.9 / 99.8, Serif 100 / 99.8, Looped 99.9 / 99.7, Ol~Chiki 100 \\
28\,px & Sans 100 / 100, Serif 100 / 100, Looped 100 / 100, Ol~Chiki 100 \\
24\,px & Sans 99.8 / 100, Serif 100 / 100, Looped 100 / 100, Ol~Chiki 100 \\
16\,px & Sans 99.5 / 99.4, Serif 99.9 / 99.0, Looped 100 / 100, Ol~Chiki 98.9 \\
\midrule
\multicolumn{2}{@{}l}{\emph{Ink of 300 Lao test lines, redrawn relative to ordinary}} \\
Ink & Sans 0.57, Serif 0.59, Looped 0.60 \\
Inked width (ordinary / redrawn) & Sans 629 / 628\,px, Serif 631 / 629\,px, Looped 627 /
625\,px \\
\midrule
\multicolumn{2}{@{}l}{\emph{Corpora of the new versions}} \\
Redrawn Lao & 818 of 149,810 lines (0.55\%) rendered identically in a redrawn font and its
original, and removed. In the test set, 1 of 1,997 lines, which is in English only \\
Santali in Devanagari & 146,275 of 149,810 lines (97.6\%) transliterate back exactly. Missing-glyph
boxes, 1.0\% of characters (3.7\% in the Ol~Chiki corpus) \\
\midrule
\multicolumn{2}{@{}l}{\emph{Santali in the pretraining text (lines with three or more of 299 Santali
word skeletons)}} \\
Calibration & Santali corpus in Devanagari, first 20,000 lines, 67.5\%. FLORES-200 Hindi, 0 of 1,012 \\
Pretraining mixture & Devanagari 0 of 266,674, Bengali 0 of 394,592, Odia 0 of 41,642 \\
Bible corpus & Devanagari 0 of 30,968, Bengali 0 of 30,964, Odia 0 of 30,968 \\
Ol~Chiki characters & in no pretraining line \\
\midrule
\multicolumn{2}{@{}l}{\emph{Ordinary Lao drawn with as little ink as redrawn Lao (ink relative to
normal renders)}} \\
Normal renders (1.00) & 9.41 / 3.41 / 14.27\% \\
Strokes lightened (0.57) & 3.02 / 1.10 / 3.20\%, lead 1.9 (1.3--2.5) \\
Strokes thinned (0.52) & 4.04 / 0.95 / 4.91\%, lead 3.1 (2.4--3.8) \\
ExtraLight font (0.49) & 1.87 / 0.80 / 1.40\%, lead 1.1 (0.6--1.5) \\
Redrawn font (0.57) & 0.27 / 0.35 / 0.35\%, difference $-0.08$ ($-0.33$ to 0.15) \\
\midrule
\multicolumn{2}{@{}l}{\emph{Santali in Devanagari without digits or shared words}} \\
FLORES-200, whole lines & 6.79 / 3.41 / 3.36\%, lead 3.4 (2.4--4.3) \\
FLORES-200, no digits & 4.08 / 2.01 / 2.08\%, lead 2.1 (1.3--2.9) \\
FLORES-200, no digits or shared words & 3.89 / 1.98 / 2.08\%, lead 1.9 (1.2--2.7) \\
IN22-Gen, whole lines & 5.96 / 2.88 / 3.22\%, lead 3.1 (2.2--4.0) \\
IN22-Gen, no digits & 3.09 / 1.53 / 2.05\% \\
IN22-Gen, no digits or shared words & 3.16 / 1.57 / 1.95\%, lead 1.6 (0.9--2.3) \\
IN22-Gen in Ol~Chiki & 500 example translations 6.17 / 4.10 / 4.20\%, gain 2.0 ($-0.1$
to 3.9). Direct 1.66 / 1.45 / 1.07\%, gain 0.6 ($-0.2$ to 1.3) \\
\midrule
\multicolumn{2}{@{}p{0.96\linewidth}@{}}{\emph{FLORES-200 sentences in the replay pool (ten runs that
replayed the FLORES-200 variant, minus the other two)}} \\
Hindi into English, gain & FLORES-200 devtest +0.5, NTREX-128 +0.8 points, difference $-0.4$ (95\%
interval $-2.4$ to 1.7) \\
\bottomrule
\end{tabular}
\end{table}

\clearpage
\normalsize
\section*{Supplementary Information}

\noindent\begin{tabular}{@{}l@{\quad}l@{}}
Supplementary Note 1 & Preliminary experiments\\
Supplementary Note 2 & Redrawn Lao and Santali in Devanagari\\
Supplementary Note 3 & Letter legibility and prior exposure\\
Supplementary Table 1 & Pretraining data by writing system and language\\
Supplementary Table 2 & Settings of rendering, pretraining and adaptation\\
Supplementary Table 3 & Every run of the Lao sweep and of Ol~Chiki adaptation
\end{tabular}

\bigskip
\noindent\textbf{Supplementary Note~1 $|$ Preliminary experiments}\par\smallskip \noindent The smaller model of Extended Data Fig.~2b has 38 million parameters and learns translation alone. It was pretrained for 60 epochs on a 23-script mixture (Methods). Each adaptation run started from the epoch-20 checkpoint of this model and trained on 48,084 lines of Lao Wikipedia with 50--70\% replay. The sweep behind the adopted objective ran eight configurations for 30 epochs each. It varied the learning rate, the mask ratio, the replay fraction and the active loss terms.

\bigskip
\noindent\textbf{Supplementary Note~2 $|$ Redrawn Lao and Santali in Devanagari}\par\smallskip \noindent\textbf{Redrawn Lao.} In all three redrawn fonts, each Lao glyph outline was replaced by a Vai outline from Noto Sans Vai (U+A500--A62B), scaled uniformly into the original bounding box. One shuffled mapping from Lao code points to Vai characters, drawn with seed~0, serves all three fonts. Contextual variants take the Vai character of their base letter, and ligatures show those of their letters side by side. The fonts keep the original advance widths, mark anchors and positioning, so line breaks, line widths and mark placement are unchanged. Lao digits are redrawn like letters, and characters outside the Lao block are unchanged.

\textbf{Santali in Devanagari.} In the transliteration table, consonants take vowel signs, and a consonant with no vowel after it takes a virama, so that the transliteration can be reversed. Aspiration produces the aspirated letter, and nasalization, length marks, digits and the two danda marks have their own signs. Of the 149,810 corpus lines, 146,275 (97.6\%) transliterate back to the identical Ol~Chiki. Latin letters, and in Ol~Chiki also ASCII punctuation, are drawn as missing-glyph boxes (Extended Data Table~4).

\bigskip
\noindent\textbf{Supplementary Note~3 $|$ Letter legibility and prior exposure}\par\smallskip \noindent\textbf{Letter legibility.} A template matcher classified 30 noisy copies of each of 52 Lao letters and 47 Ol~Chiki letters at 44, 28, 24 and 16 pixels. Lao marks, which cannot stand alone, were drawn on a carrier letter. In ordinary and in redrawn Lao, the matcher identified each letter with at least 99.0\% accuracy at every size, and Ol~Chiki letters with at least 98.9\% (Extended Data Table~4).

\textbf{Prior exposure.} We reduced the 500 most frequent Santali words of three or more letters in the Ol~Chiki corpus, transliterated into Devanagari, and all pretraining words in Devanagari, Bengali and Odia to a spelling skeleton. It drops vowel length, nasal marks, visarga, virama and nukta, writes va as ba and maps Bengali and Odia onto Devanagari. We kept the 299 skeletons of four or more characters that are absent from the FLORES-200 devtest texts of Hindi, Marathi, Nepali, Bengali, Assamese and Odia. A line passes the screen if it contains three or more different ones. The screen passes 67.5\% of the first 20,000 Devanagari lines of the Santali corpus, and none of the 1,012 FLORES-200 Hindi sentences. Extended Data Table~4 gives the counts for the Devanagari, Bengali and Odia lines of the pretraining mixture and of the Bible corpus.

\clearpage
\footnotesize
\begin{longtable}{@{}p{0.19\linewidth}p{0.47\linewidth}rr@{}}
\caption*{\normalsize\textbf{Supplementary Table~1 $|$ Pretraining data by writing system and language.} The pretraining mixture holds 4,113,720 meaning groups and 9,774,015 lines in 27 writing systems. Sentences come from OPUS, with English as the pivot language, and from the eBible corpus of verse-aligned Bible translations in native scripts, one translation per script, the one with the most verses. Lines give the number of sentences in each writing system in the mixture, as plotted in Fig.~2b, and Bible how many of them come from the eBible corpus. A Bible translation is in the language of the OPUS text of its script unless a second language is named. Two of them, in Arabic and Ethiopic script, are in languages other than Arabic and Amharic. The Latin-script lines are the English side of OPUS. Each script is drawn in one Noto font family.}\\
\toprule
Writing system & Language & Lines & Bible \\
\midrule
\endfirsthead
\toprule
Writing system & Language & Lines & Bible \\
\midrule
\endhead
\bottomrule
\endfoot
Latin & English & 4,077,098 & 0 \\
Cyrillic & Russian; Ukrainian (Bible) & 491,966 & 30,959 \\
Greek & Greek; Koine Greek (Bible) & 475,873 & 26,415 \\
Armenian & Armenian & 4,690 & 0 \\
Georgian & Georgian & 205,687 & 0 \\
Hebrew & Hebrew & 477,980 & 30,966 \\
Arabic & Arabic; another language (Bible) & 489,340 & 31,153 \\
Syriac & Assyrian Neo-Aramaic & 10,418 & 10,418 \\
Ethiopic & Amharic; another language (Bible) & 79,289 & 30,371 \\
Devanagari & Hindi & 266,674 & 30,968 \\
Bengali & Bengali & 394,592 & 30,964 \\
Gurmukhi & Punjabi & 84,685 & 30,968 \\
Gujarati & Gujarati & 111,336 & 30,968 \\
Odia & Odia & 41,642 & 30,968 \\
Tamil & Tamil & 175,771 & 30,967 \\
Telugu & Telugu & 62,682 & 30,868 \\
Kannada & Kannada & 41,882 & 30,955 \\
Malayalam & Malayalam & 382,982 & 30,964 \\
Sinhala & Sinhala; Sanskrit (Bible) & 363,624 & 7,957 \\
Tibetan & Tibetan & 30,917 & 30,917 \\
Limbu & Limbu & 7,873 & 7,873 \\
Thai & Thai & 468,940 & 30,966 \\
Khmer & Khmer; Sanskrit (Bible) & 84,005 & 7,957 \\
Myanmar & Burmese & 36,534 & 30,909 \\
Han & Chinese & 464,164 & 7,940 \\
Hangul & Korean & 435,433 & 0 \\
Kana and kanji & Japanese & 7,938 & 7,938 \\
\midrule
Total & & 9,774,015 & 571,329 \\
\end{longtable}

\clearpage
\begin{longtable}{@{}p{0.30\linewidth}p{0.66\linewidth}@{}}
\caption*{\normalsize\textbf{Supplementary Table~2 $|$ Settings of rendering, pretraining and adaptation.} The settings of every model in the paper, as described in Methods. Supplementary Note~1 describes the preliminary model used to choose the adaptation objective.}\\
\toprule
Setting & Value \\
\midrule
\endfirsthead
\toprule
Setting & Value \\
\midrule
\endhead
\bottomrule
\endfoot
\multicolumn{2}{@{}l}{\emph{Rendering}} \\
Image & 64 by 768 pixels, greyscale \\
Font size & starts at 0.7 times the image height and shrinks until the text fits, down to 0.44 times
the height, about 28 pixels \\
Text that does not fit & cut at a word or character boundary \\
Layout & RAQM shaping where available \\
Augmentation in training & shift of up to 2 pixels, rotation of up to 2 degrees, Gaussian noise
\\
Evaluation renders & clean and deterministic \\
\midrule
\multicolumn{2}{@{}l}{\emph{Architecture}} \\
Input & 8 by 8-pixel patches with fixed two-dimensional sinusoidal position encodings \\
Blocks & layer normalization before each sublayer, feed-forward layers four times the width \\
Output & mean pooling after a final layer normalization, then a two-layer projection head
to an L2-normalized embedding \\
GlyphMind-205M & width 1,024, 16 blocks, 16 attention heads, embedding size 128, 204.9 million
parameters \\
GlyphMind-38M & width 512, 12 blocks, 8 attention heads, embedding size 64 \\
\midrule
\multicolumn{2}{@{}l}{\emph{Pretraining objective}} \\
Contrastive loss & InfoNCE over positives within each GPU's batch (32 records for GlyphMind-205M, 64 for GlyphMind-38M), temperature 0.07 \\
Positives & other members of the meaning group, in other scripts, one drawn at random at each step. A Bible record keeps up to eight \\
Cipher positive & with probability 0.3, a substitution cipher of the group's Latin-script text
with new letter shapes, in records with an English anchor \\
Margin loss & margin 0.2 against hard negatives, weight 1.0, one hard negative per example, drawn at random from the record \\
Hard negatives & up to three edits of one sentence of the group, rendered like it: letters shuffled inside each word of four or more characters, keeping its first and last letter (in runs of three characters in text without spaces); one such word swapped for a random English word, in English text only; digits changed. OPUS records edit their English anchor, and all records of a Bible verse share the edits of one of its translations \\
Predictive term & GlyphMind-205M only. Masked spans of 3--8 full-height patch columns, about 50\% of
the ink. Targets from an exponential-moving-average teacher with momentum 0.999. Weight 1.0 \\
\midrule
\multicolumn{2}{@{}l}{\emph{Pretraining of GlyphMind-205M}} \\
Optimizer & AdamW with $\beta_1=0.9$ and $\beta_2=0.95$, weight decay 0.05, gradient clipping at norm 1.0, bfloat16 precision \\
Learning rate & linear warm-up to $3{\times}10^{-4}$ over the first three epochs (5\% of the steps of the 60-epoch schedule), then a cosine schedule that would reach zero at epoch 60; $8.2{\times}10^{-5}$ at the final
checkpoint \\
Batch & global batch of 256 on eight A100 GPUs, 32 per GPU \\
Length & 40 epochs of the 60-epoch schedule, checkpoints every five epochs \\
\midrule
\multicolumn{2}{@{}l}{\emph{Pretraining of the two GlyphMind-38M models}} \\
Objective & as above without the predictive term \\
Optimizer & as for GlyphMind-205M. Linear warm-up to $3{\times}10^{-4}$ over the first 5\% of steps, then a cosine schedule to zero \\
Batch & global batch of 256 on four A100 GPUs, 64 per GPU \\
Length & 30 epochs. After epoch 10, both runs continued with a fresh optimizer state and a
new 20-epoch cosine schedule \\
\midrule
\multicolumn{2}{@{}l}{\emph{Adaptation of GlyphMind-205M}} \\
Objective on new lines & masked spans and two sentence-level targets, the pooled feature and the
projection output, each with weight 1.0. Mask ratio 0.5. The teacher is a frozen copy of the starting checkpoint. \\
Replay & translation pairs from a pretraining mixture, 50\% or 70\% of training. Each example is a replayed record with that probability and otherwise a new line, so every batch mixes the two, and an epoch has as many examples as the new corpus has lines. Replayed records keep the contrastive and margin losses of pretraining, without cipher positives, computed among the replayed records of the batch. The two parts of the loss are added with equal weight \\
Learning rate & constant, $1$, $3$ or $5{\times}10^{-5}$, after a linear warm-up over the first 3\% of steps \\
Optimizer and batch & AdamW as in pretraining, batch of 32 on one A100 GPU \\
Checkpoints & every three epochs in the Lao sweep, every ten epochs in the four-version
experiment \\
Lao sweep & 12 epochs. Two configurations ran 12 more epochs after a reset of the optimizer state
(Supplementary Table~3) \\
Four-version experiment & epoch-40 checkpoint, learning rate $3{\times}10^{-5}$, replay 50\%, 80
epochs without a reset, three seeds per version \\
\end{longtable}

\clearpage
\begin{longtable}{@{}p{0.27\linewidth}cccc>{\raggedleft\arraybackslash}p{0.07\linewidth}>{\raggedleft\arraybackslash}p{0.07\linewidth}@{}}
\caption*{\normalsize\textbf{Supplementary Table~3 $|$ Every run of the Lao sweep and of Ol~Chiki adaptation.} Top-5 retrieval into English at the final checkpoint, in per cent. Lao runs are scored on NTREX-128 Lao and Ol~Chiki runs on FLORES-200 devtest Santali in Ol~Chiki, both on pure renders. ex, with 500 example translations. dir, direct retrieval. The Lao sweep varied the initial checkpoint (epoch 20 or 40), the learning rate ($1$, $3$ or $5{\times}10^{-5}$) and the replay fraction (0.5 or 0.7). All runs in this table replayed the pretraining mixture. The two extensions continue configurations 6 and 8. ex averages three random draws of the 500 example translations, and ten for the two runs that belong to the four-version experiment, setting 1 of Ol~Chiki and the subsampled Lao run (Methods). The strongest simple baseline reaches 2.05\% for Lao and 3.36\% for Santali in direct retrieval.}\\
\toprule
Run & Initial epoch & Learning rate & Replay & Epochs & ex & dir \\
\midrule
\endfirsthead
\toprule
Run & Initial epoch & Learning rate & Replay & Epochs & ex & dir \\
\midrule
\endhead
\bottomrule
\endfoot
\multicolumn{7}{@{}l}{\emph{Lao sweep, 991,775 lines}} \\
Configuration 1 & 20 & $1{\times}10^{-5}$ & 0.7 & 12 & 11.08 & 4.91 \\
Configuration 2 & 40 & $1{\times}10^{-5}$ & 0.7 & 12 & 10.62 & 6.21 \\
Configuration 3 & 20 & $3{\times}10^{-5}$ & 0.7 & 12 & 14.50 & 7.46 \\
Configuration 4 & 40 & $3{\times}10^{-5}$ & 0.7 & 12 & 14.13 & 8.21 \\
Configuration 5 & 20 & $5{\times}10^{-5}$ & 0.7 & 12 & 15.33 & 7.36 \\
Configuration 6 & 40 & $5{\times}10^{-5}$ & 0.7 & 12 & 16.21 & 8.71 \\
Configuration 7 & 20 & $3{\times}10^{-5}$ & 0.5 & 12 & 13.92 & 7.21 \\
Configuration 8 & 40 & $3{\times}10^{-5}$ & 0.5 & 12 & 15.96 & 10.07 \\
Reported run, after the reset & 40 & $5{\times}10^{-5}$ & 0.7 & 12 + 12 & 21.25 & 14.27 \\
Second extension & 40 & $3{\times}10^{-5}$ & 0.5 & 12 + 12 & 18.96 & 12.82 \\
\midrule
\multicolumn{7}{@{}l}{\emph{Ol~Chiki runs, 149,810 lines, replay 0.5}} \\
Setting 1 & 40 & $3{\times}10^{-5}$ & 0.5 & 80 & 5.78 & 1.09 \\
Setting 2 & 40 & $5{\times}10^{-5}$ & 0.5 & 80 & 7.87 & 0.69 \\
Setting 3 & 20 & $3{\times}10^{-5}$ & 0.5 & 80 & 7.93 & 0.99 \\
Extension of setting 2 & & & & 80 + 12 & 7.13 & 1.68 \\
Extension of setting 3 & & & & 80 + 12 & 7.33 & 1.09 \\
\multicolumn{5}{@{}l}{Highest direct retrieval in any of these runs, at any checkpoint} & & 1.78 \\
\midrule
\multicolumn{7}{@{}l}{\emph{Subsampled Lao, 149,810 lines, NTREX-128}} \\
Setting 1 of Ol~Chiki & 40 & $3{\times}10^{-5}$ & 0.5 & 80 & 16.40 & 9.46 \\
\end{longtable}
\end{document}